\documentclass[11pt]{article}

\usepackage[utf8]{inputenc} 
\usepackage[T1]{fontenc}    
\usepackage{hyperref}       
\usepackage{url}            
\usepackage[numbers,sort&compress]{natbib} 
\usepackage{booktabs}       
\usepackage{amsfonts}       
\usepackage{amsmath}        
\usepackage{nicefrac}       
\usepackage{fancyhdr}       
\usepackage{graphicx}       
\usepackage{algorithm}      
\usepackage{algorithmic}    
\usepackage{geometry}       
\usepackage{titlesec}       

\usepackage{float}          
\usepackage{placeins}
\usepackage[font=small]{caption}
\usepackage{etoolbox} 

\graphicspath{{./}{./images/}{./figures/}{./assets/}}     

\fancypagestyle{firstpage}{
    \fancyhf{}
    \fancyhead[L]{%
        \includegraphics[height=1.0cm,keepaspectratio]{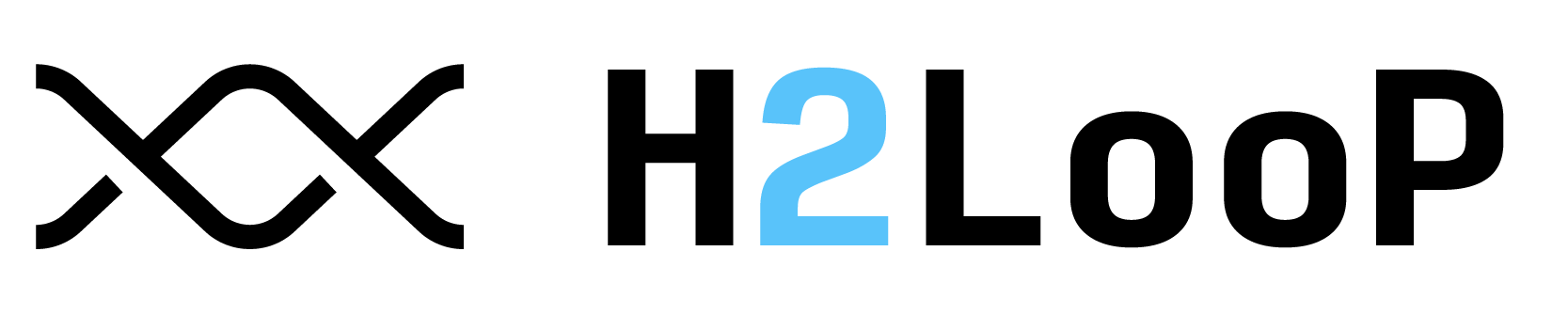}
    }
    
}

\makeatletter
\renewcommand{\maketitle}{
  \begin{center}
    {\LARGE\bfseries \@title \par}
    \vskip 1.5em
    {\large \@author \par}
  \end{center}
  \vskip 2em
}
\makeatother

\titleformat{\section}
  {\Large\bfseries}{\thesection}{1em}{}
\titleformat{\subsection}
  {\large\bfseries}{\thesubsection}{1em}{}
\titleformat{\subsubsection}
  {\normalsize\bfseries}{\thesubsubsection}{1em}{}

\title{H2LooP Telecom Model v1: From Telecom Comprehension to Autonomous Issue and PR Resolution}

\author{
\begin{tabular}{c c c c c}
  Amit Singh & Vedant Nipane & Mayank Goel & Pulkit Agrawal & Sairanjan Mishra \\
\end{tabular}
\\[0.5em]
\textbf{H2LooP.ai}
}

\begin{document}
\thispagestyle{firstpage}

\maketitle

\begin{abstract}
We present H2LooP Telecom Model v1, a domain-specialized large language model fine-tuned for the telecommunications industry. We release two domain-adapted model variants serving complementary use cases: a comprehension-focused variant for telecom domain question answering and reasoning, and an agentic variant for autonomous telecom code generation, pull request resolution, and code commits on production repositories. H2LooP Telecom achieves strong results on the GSMA Open Telecom Lite (OT-Lite) benchmark and a proprietary telecom code generation benchmark, outperforming frontier closed-source models such as GPT-5 and Claude Opus on independent leaderboard evaluation, while preserving general-purpose capabilities. The Comprehension variant achieves 81.8\% weighted average on OT-Lite Pass@3, and, independently, ranks 5th overall on the official community-run Open Telco AI Leaderboard\textsuperscript{*} at only 31B parameters-ahead of frontier closed-source systems including Claude Opus 4.6, GPT-5, Gemini 3 Flash, Grok-4-fast, and Kimi K2.5. Our agentic variant obtains a relative improvement of +8.8\% in AST Similarity and +20.0\% in Location IoU over the base model on telecom code generation, while maintaining identical MMLU (74.0\%) and BFCL v3 multi-turn function calling (79.0\%) performance, indicating zero catastrophic forgetting. Domain specialization on curated telecom corpora, spanning 3GPP standards, O-RAN specifications, network telemetry, and real repository commits, yields substantial improvements over general-purpose models of equivalent scale, approaches frontier closed-source models on domain-specific evaluation, and is independently corroborated by our official leaderboard standing.
\end{abstract}

\section{Introduction}

The telecommunications industry is undergoing a fundamental transformation driven by software-defined networking, Open RAN disaggregation, and the increasing complexity of 5G and 6G standards. Modern telecom infrastructure relies on millions of lines of code spanning protocol stacks, network functions, and orchestration systems. However, general-purpose large language models (LLMs), despite their broad capabilities, exhibit insufficient performance on tasks requiring specialized knowledge of telecom standards, network log diagnosis, and protocol-conformant code generation.

This capability gap creates a significant bottleneck in telecom software development. Engineers must manually navigate thousands of pages of 3GPP specifications, interpret proprietary log formats, and maintain codebases that implement continuously evolving standards. An AI system with deep domain knowledge could substantially accelerate development velocity across these tasks.

We address this gap with H2LooP Telecom Model v1, a model specialized for telecommunications through supervised fine-tuning on curated domain-specific datasets. Our key contributions are as follows:

\begin{enumerate}
    \item \textbf{Domain-specialized training corpora} comprising non-coding instruction-response pairs covering telecom standards, specifications, mathematical reasoning, log analysis, and structured data interpretation, alongside a coding corpus derived from real telecom repository commits.

    \item \textbf{Two model variants} serving complementary deployment scenarios: H2LooP Telecom Comprehension for domain question answering and reasoning, and H2LooP Telecom Agent for autonomous code generation and pull request resolution.

    \item \textbf{Strong OT-Lite performance}, with 81.8\% weighted average (Comprehension variant) surpassing Qwen3.6-27B~\cite{qwen3} (+1.8~pp) and Nemotron-30B~\cite{nemotron} (+12.7~pp), while exceeding the frontier-class GLM-5~\cite{glm5} (744B, 40B active) by +2.5~pp and approaching Gemini 3.1 Pro~\cite{gemini} (84.2\%).

    \item \textbf{Autonomous PR resolution} validated on 11 production telecom repositories spanning RAN, core network, simulation, and research codebases.

    \item \textbf{Zero catastrophic forgetting} as evidenced by identical MMLU performance (74.0\%) between the fine-tuned Agent variant and the base model.

    \item \textbf{Independent third-party verification}\textsuperscript{*}: on the community-run Open Telco AI Leaderboard, H2LooP Telecom Comprehension ranks 5th overall at only 31B parameters, outperforming frontier closed-source systems including Claude Opus 4.6~\cite{claude}, GPT-5~\cite{gpt5}, Gemini 3 Flash~\cite{gemini}, Grok-4-fast~\cite{grok4}, and Kimi K2.5~\cite{kimi} (Section~\ref{sec:leaderboard}).
\end{enumerate}

\section{Related Work}

\subsection{Domain-Specialized Language Models}

Domain specialization of large language models has been investigated across medicine~\cite{singhal2023}, law~\cite{colombo2024}, and finance~\cite{wu2023}, establishing that targeted fine-tuning on domain corpora yields substantial improvements on in-domain evaluation. In the telecommunications domain, AdaptKey Nemotron-30B~\cite{nemotron} applies architectural modifications (key-value attention adaptation) for telecom specialization but exhibits significant capability degradation on coding and comprehension benchmarks relative to models of comparable parameter count.

\subsection{Telecom Benchmarks}

The GSMA Open Telecom (OT) benchmark suite~\cite{otbench} provides standardized evaluation across telecom comprehension dimensions including standards knowledge, network log interpretation, mathematical reasoning, and structured data understanding. OT-Lite~\cite{otlite}, its lightweight variant, enables efficient evaluation with Pass@3 scoring across 10 sub-benchmarks covering distinct telecom competency areas. We adopt OT-Lite as our primary comprehension evaluation protocol.

Two sub-benchmarks within the OT-Lite suite merit particular attention. \textbf{teleqna-v2}, drawn from the SuLLMerica/TeleQnA\_Train\_With\_RAG\_Context dataset (HuggingFace), addresses a gap in the original TeleQnA: whereas TeleQnA skews toward general telecom knowledge (definitions, architecture concepts), teleqna-v2 tests deep 3GPP Release 17/18 specifics-sidelink/V2X, NR carrier aggregation edge cases, 5GC charging procedures, SDT, and RF emissions limits. It comprises approximately 82\% standards specification questions and 18\% standards overview questions, evaluating whether a model actually knows the spec text rather than merely the domain. \textbf{telereason} is the only open-ended benchmark in the suite, requiring free-form answers evaluated on specific technical criteria rather than multiple-choice selection. It spans four domains: 5GC interface interactions ($\sim$32\%), alarm and fault diagnosis ($\sim$27\%), RAN procedures ($\sim$27\%), and protocol procedures ($\sim$14\%). This makes it the most discriminative benchmark in the suite, with scores varying widely across models compared to the tight bands typical of MCQ benchmarks.

\subsection{Code Generation for Telecom}

General code generation benchmarks such as HumanEval~\cite{humaneval} and SWE-bench~\cite{swebench} evaluate broad programming ability but do not capture the unique challenges inherent to telecom codebases: protocol state machines, ASN.1-derived data structures, RRC/NAS message handling, FAPI interfaces, and hardware-abstraction layers. We introduce a telecom-specific coding benchmark derived from real commit histories of open-source telecom projects, evaluated across four complementary metrics.

\subsection{Agentic Code Models}

Recent work on autonomous coding agents, including SWE-Agent~\cite{sweagent} and Devin~\cite{devin}, demonstrates the viability of LLM-driven pull request generation. These systems rely on general-purpose models that lack telecom domain knowledge, fundamentally limiting their effectiveness on specialized codebases where understanding of protocol semantics, standards compliance, and domain-specific APIs is required for correct code generation. H2LooP Telecom Agent unifies domain comprehension with agentic coding capability within a single model.

\section{Data Strategy}

We construct two complementary training datasets designed to maximize coverage of telecom knowledge and coding patterns. Our methodology emphasizes dataset quality, domain authenticity, and distributional coverage over scale alone.

\subsection{Non-Coding Corpus}

Our non-coding dataset comprises instruction-response pairs distributed across the following telecom domains:

\textbf{3GPP Standards and Specifications.} Questions and explanations drawn from Release 15 through Release 18 documents covering New Radio (NR), Long Term Evolution (LTE), core network architecture, and protocol procedures.

\textbf{O-RAN Architecture.} Coverage of O-RAN Alliance specifications including the RAN Intelligent Controller (RIC), fronthaul interface splits, and disaggregated network function interactions.

\textbf{Network Telemetry and Logs.} Interpretation of telecom-specific log formats, Key Performance Indicator (KPI) analysis, alarm correlation, and root cause diagnosis patterns.

\textbf{Telecom Mathematics.} Signal processing fundamentals, channel modeling, link budget calculations, resource allocation optimization, and information-theoretic computations.

\textbf{Structured Data.} Table interpretation from standards documents, specification cross-referencing, and parameter relationship reasoning across multi-table contexts.

\textbf{6G Research.} Emerging concepts including Reconfigurable Intelligent Surfaces (RIS), terahertz communications, AI-native network architectures, and joint communication and sensing.

\begin{figure}[H]
    \centering
    \includegraphics[width=0.75\linewidth]{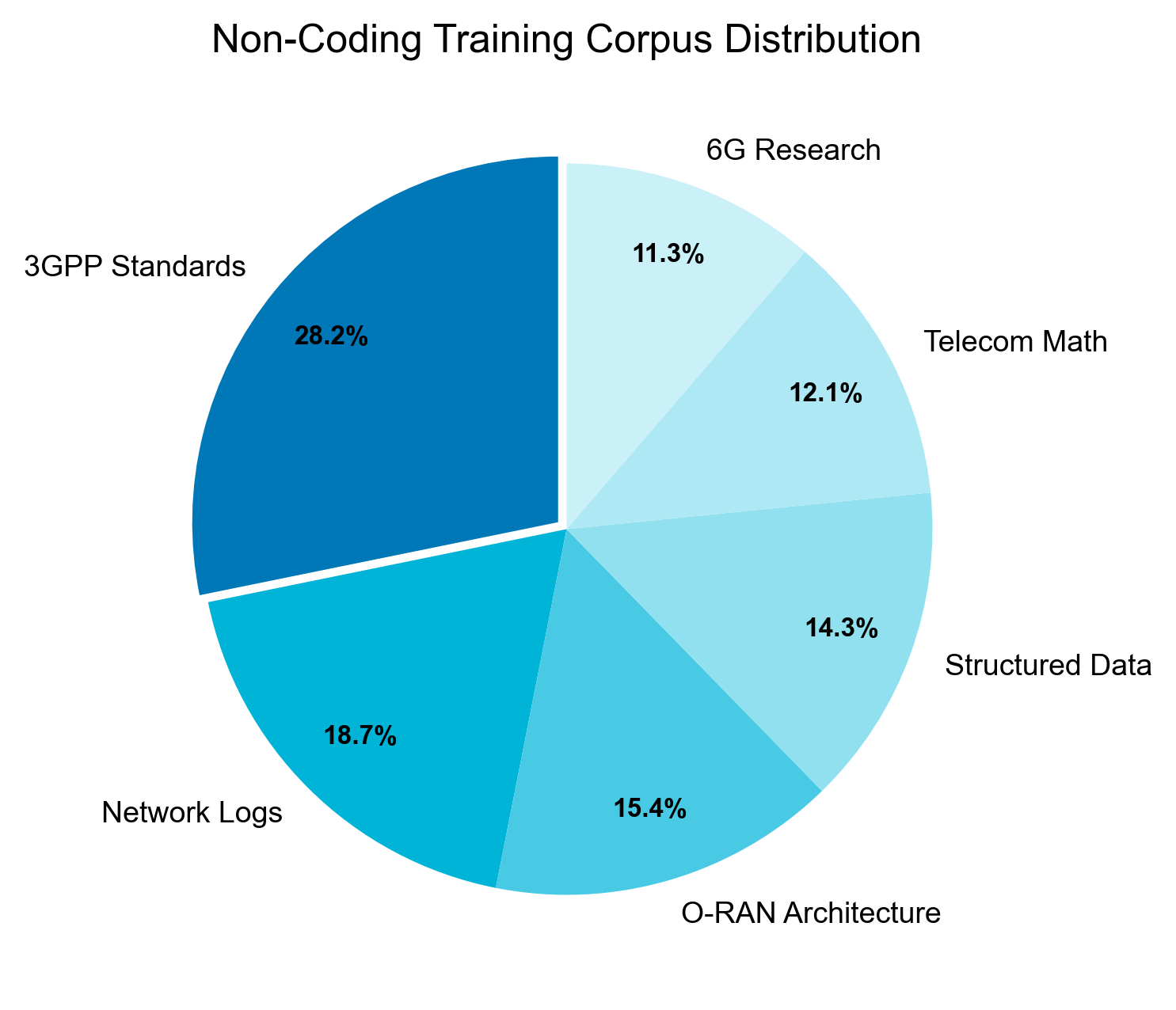}
    \caption{Non-coding training corpus domain distribution.}
    \label{fig:corpus_distribution}
\end{figure}

The corpus is constructed to reflect the empirical distribution of tasks encountered by telecom engineers in practice, with emphasis on multi-step reasoning over simple factual recall.

\subsection{Coding Corpus}

Our coding dataset is derived from real commit histories of open-source telecom repositories. Each training sample encodes a code change contextualized by:

\begin{itemize}
    \item The repository structure and relevant surrounding code context
    \item The issue description or intent motivating the change
    \item The precise file locations and edits constituting the commit diff
\end{itemize}

This format trains the model to perform end-to-end pull request resolution: given a natural language description of a required change, the model must identify the correct files, locate the relevant code regions, and generate syntactically and semantically correct edits conforming to project conventions. The training repositories span major open-source telecom projects covering RAN implementations, core network functions, and simulation environments.

\begin{figure}[H]
    \centering
    \includegraphics[width=0.9\linewidth]{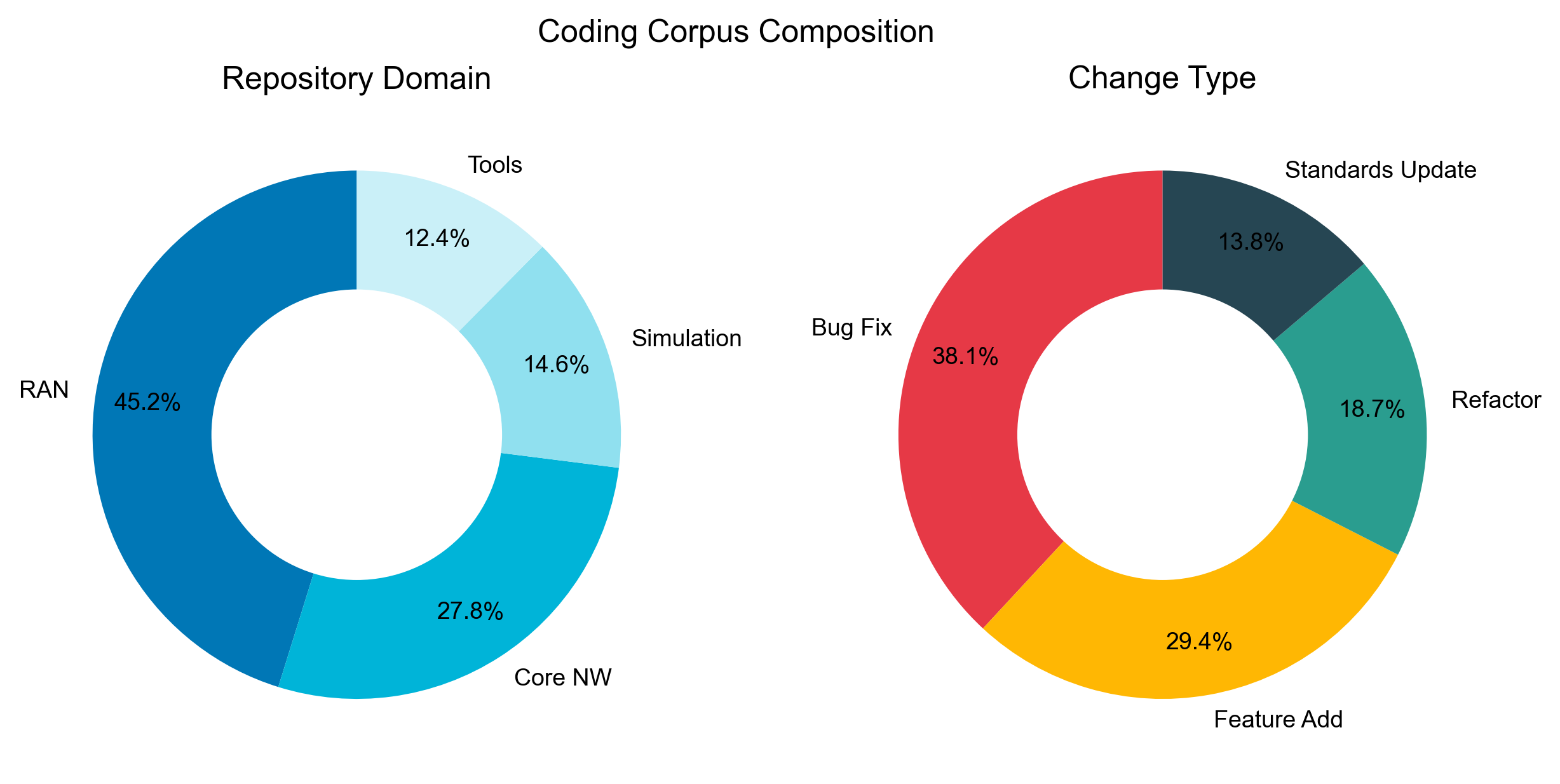}
    \caption{Coding corpus composition-repository domain (left) and change type (right).}
    \label{fig:coding_corpus}
\end{figure}

\section{Model Architecture and Variants}

Both model variants are derived from a dense transformer base model via supervised fine-tuning (SFT) with Low-Rank Adaptation (LoRA). We do not modify the base model architecture or attention mechanism, preserving compatibility with existing inference infrastructure.

\subsection{H2LooP Telecom Comprehension}

The Comprehension variant is optimized exclusively for domain question answering and reasoning. It achieves the highest OT-Lite scores among our releases (81.8\% weighted average) and is intended for non-agentic deployment scenarios including standards consultation, log analysis, specification interpretation, and telecom education. This variant does not incorporate function calling or agentic training objectives.

\subsection{H2LooP Telecom Agent}

The Agent variant constitutes our primary release, optimized for the complete pipeline from telecom comprehension to autonomous code generation. The training objective balances domain knowledge acquisition with three agentic capabilities:

\begin{itemize}
    \item \textbf{Multi-turn function calling} for tool-augmented agentic workflows
    \item \textbf{Telecom code generation} specialized for domain-specific repositories
    \item \textbf{PR resolution} encompassing file localization, edit generation, and commit structuring
\end{itemize}

The Agent variant maintains strong comprehension performance (78.8\% OT-Lite weighted average) while achieving superior coding metrics and preserving multi-turn function calling capability at parity with the base model.

\section{Experimental Setup}

\subsection{Evaluation Protocol}

All evaluations employ greedy decoding (temperature = 0.0) unless otherwise specified. For stochastic evaluations (OT-Lite Pass@3), we use temperature = 0.7 with three independent samples per query, where a question is considered passed if any of the three attempts is correct. Models are served via an OpenAI-compatible API with a 64K token context window. Function calling evaluations use the native tool calling format.

\subsection{Baselines}

We compare against models in two categories:

\textbf{Similarly-sized open-weight models (27B--35B parameters):} Qwen3.6-27B~\cite{qwen3}, Qwen3.6-35B-A3B~\cite{qwen3}, Gemma4-26B-MoE, Nemotron-30B~\cite{nemotron}, and AdaptKey Nemotron-30B~\cite{nemotron}.

\textbf{Frontier models (reference ceiling):} Gemini 3.1 Pro~\cite{gemini}, Gemini 3 Flash~\cite{gemini}, GLM-5~\cite{glm5}.

\textbf{Additional large-scale reference:} Qwen3-480B~\cite{qwen3} (included to demonstrate specialization efficiency relative to scale).

\subsection{Evaluation Benchmarks}

\begin{table}[H]
\centering
\caption{Summary of Evaluation Benchmarks.}
\label{tab:eval_benchmarks}
\resizebox{\textwidth}{!}{%
\begin{tabular}{llll}
\toprule
\textbf{Benchmark} & \textbf{Capability Tested} & \textbf{Format} & \textbf{Scoring} \\
\midrule
OT-Lite~\cite{otlite} & Telecom domain comprehension & MCQ + open-ended (10 sub-benchmarks) & Pass@3 weighted average \\
Telecom Coding & Code generation on telecom repos & Commit-level patch generation & BLEU-4, AST Sim, Edit-Dist, Location IoU \\
BFCL v3~\cite{bfcl} & Multi-turn function calling & Tool calling via OpenAI API & Accuracy across 4 categories \\
MMLU~\cite{mmlu} & General knowledge retention & MCQ across non-telecom subjects & Exact match \\
\bottomrule
\end{tabular}%
}
\end{table}

\section{Results}

\subsection{Telecom Comprehension: OT-Lite Pass@3}

We evaluate on the GSMA Open Telecom Lite benchmark~\cite{otlite} using the Pass@3 protocol across all 10 sub-benchmarks. Each sub-benchmark targets a distinct telecom competency area, and the final score is computed as the weighted average across all sub-benchmarks following the official GSMA weighting scheme. All baseline models reported below reflect the versions publicly available at the time of measurement; model availability and reported capabilities are subject to change as vendors release updates.

\begin{table}[H]
\centering
\caption{OT-Lite Pass@3 Weighted Average (All Models)\textsuperscript{\ddag}.}
\label{tab:otlite_ranking}
\resizebox{\textwidth}{!}{%
\begin{tabular}{llcc}
\toprule
\textbf{Model} & \textbf{Parameters} & \textbf{Weighted Avg} & \textbf{$\Delta$ vs.\ H2LooP Compr.} \\
\midrule
Gemini 3.1 Pro~\cite{gemini} & Undisclosed & 84.2\% & +2.4~pp \\
Gemini 3 Flash~\cite{gemini} & Undisclosed & 83.8\% & +2.0~pp \\
\textbf{H2LooP Telecom Comprehension} & 31B & \textbf{81.8\%} & -- \\
Qwen3.6-27B~\cite{qwen3} & 27B & 80.0\% & $-$1.8~pp \\
Qwen3.6-35B-A3B~\cite{qwen3} & 35B (3B active) & 79.9\% & $-$1.9~pp \\
GLM-5~\cite{glm5} & 744B (40B active) & 79.3\% & $-$2.5~pp \\
Gemma4-26B-MoE~\cite{gemma} & 26B & 78.0\% & $-$3.8~pp \\
Qwen3-480B~\cite{qwen3} & 480B & 72.9\% & $-$8.9~pp \\
Nemotron-30B~\cite{nemotron} & 30B & 69.1\% & $-$12.7~pp \\
AdaptKey Nemotron-30B~\cite{nemotron} & 30B & 68.9\% & $-$12.9~pp \\
\bottomrule
\end{tabular}%
}
\end{table}

\begin{figure}[H]
    \centering
    \includegraphics[width=0.9\linewidth]{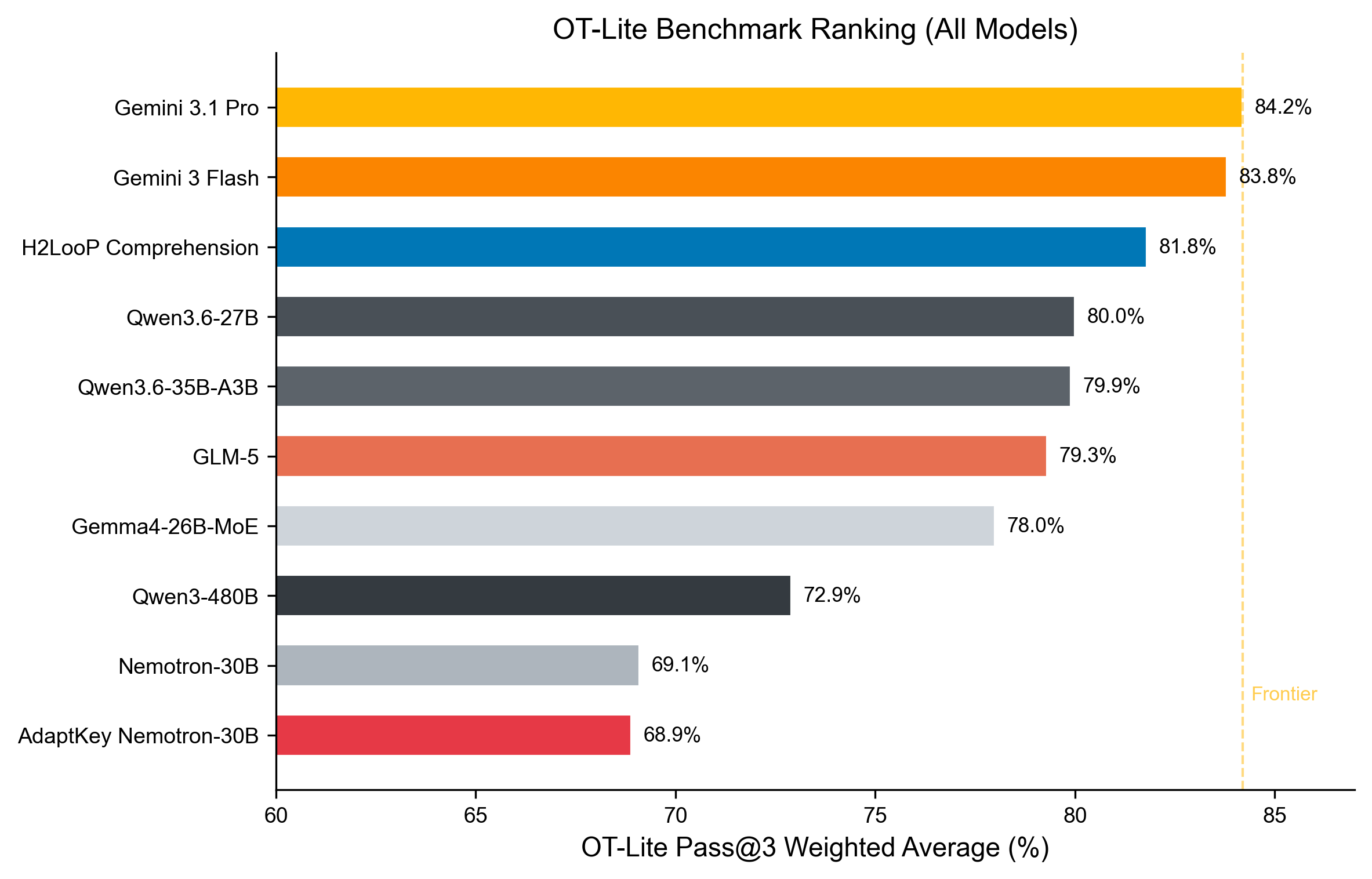}
    \caption{OT-Lite Pass@3 weighted average ranking across all models.}
    \label{fig:otlite_ranking}
\end{figure}

H2LooP Telecom Comprehension surpasses the open-weight baselines evaluated above (Qwen3.6-27B, Qwen3.6-35B-A3B, GLM-5, Gemma4-26B-MoE, Nemotron-30B, and AdaptKey Nemotron-30B) and approaches frontier closed-source systems, trailing Gemini 3 Flash by only 2.0 percentage points. Notably, our model surpasses Qwen3-480B (72.9\%), a model with substantially more parameters, by 8.9~pp. This result demonstrates that domain-specialized fine-tuning is substantially more parameter-efficient than general-purpose scaling for telecom-specific tasks.

\begin{table}[H]
\centering
\caption{OT-Lite Sub-benchmark Breakdown.}
\label{tab:otlite_subbench}
\resizebox{\textwidth}{!}{%
\begin{tabular}{lcccccc}
\toprule
\textbf{Sub-benchmark} & \textbf{H2LooP Compr. (31B)} & \textbf{Qwen3.6-27B} & \textbf{GLM-5} & \textbf{Nemotron-30B} & \textbf{AdaptKey} & \textbf{TSLAM-150B~\cite{tslam}\textsuperscript{\dag}} \\
\midrule
srsranbench & 90.8\% & 90.7\% & 88.7\% & 80.0\% & 80.7\% & 80.0\% \\
oranbench & 88.9\% & 86.7\% & 86.0\% & 73.3\% & 81.3\% & 80.7\% \\
three-gpp & 50.4\% & 48.0\% & 62.0\% & 34.0\% & 28.0\% & 80.0\% \\
teletables & 60.0\% & 53.0\% & 47.0\% & 40.0\% & 44.0\% & 70.0\% \\
telemath & 80.5\% & 82.0\% & 81.0\% & 64.0\% & 46.0\% & 71.0\% \\
telelogs & 80.3\% & 58.0\% & 50.0\% & 27.0\% & 29.0\% & 39.0\% \\
sixg-bench & 75.5\% & 90.0\% & 88.0\% & 75.3\% & 81.3\% & -- \\
teleqna & 86.6\% & 84.9\% & 83.8\% & 79.1\% & 78.2\% & 80.6\% \\
teleqna-v2 & 86.2\% & 80.0\% & 82.5\% & 69.5\% & 74.5\% & -- \\
telereason & 51.3\% & 56.0\% & 58.0\% & 28.0\% & 6.0\% & -- \\
\bottomrule
\end{tabular}%
}
\end{table}

\begin{figure}[H]
    \centering
    \includegraphics[width=0.8\linewidth]{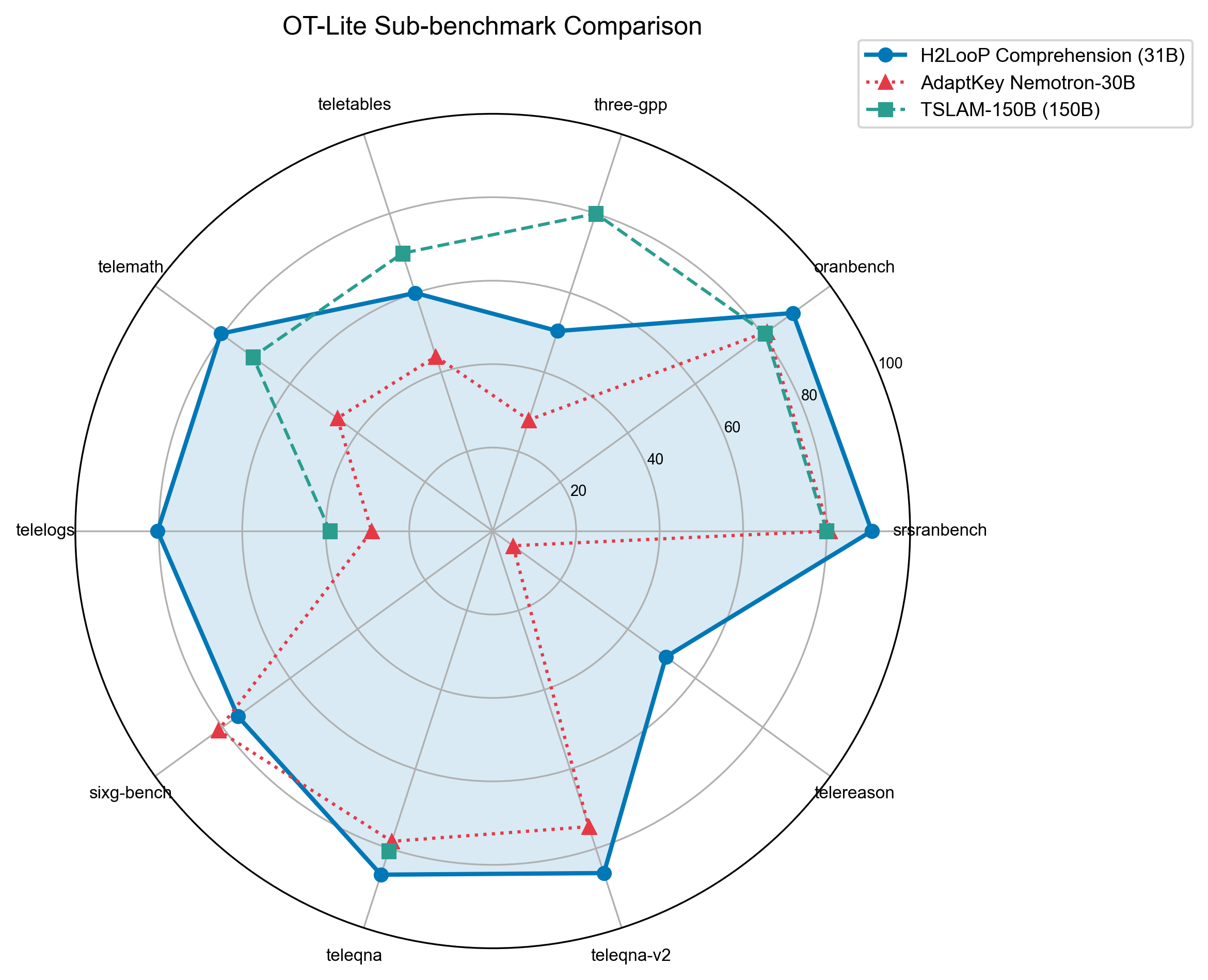}
    \caption{OT-Lite sub-benchmark radar comparison, including TSLAM-150B\textsuperscript{\dag} on the 7 sub-benchmarks it was evaluated on.}
    \label{fig:radar_otlite}
\end{figure}

Despite having roughly 5$\times$ fewer parameters (31B vs.\ 150B), H2LooP Telecom Comprehension outperforms TSLAM-150B on 5 of the 7 shared sub-benchmarks-most notably telelogs (80.3\% vs.\ 39.0\%, +41.3~pp) and oranbench (88.9\% vs.\ 80.7\%, +8.2~pp)-and achieves a higher average across that shared subset (76.8\% vs.\ 71.6\%, +5.2~pp). TSLAM-150B leads on three-gpp (80.0\% vs.\ 50.4\%) and teletables (70.0\% vs.\ 60.0\%), but its substantially larger parameter count does not translate into a consistent advantage, reinforcing the parameter-efficiency argument developed in Section~\ref{sec:parameter_efficiency}.

\begin{figure}[H]
    \centering
    \includegraphics[width=0.8\linewidth]{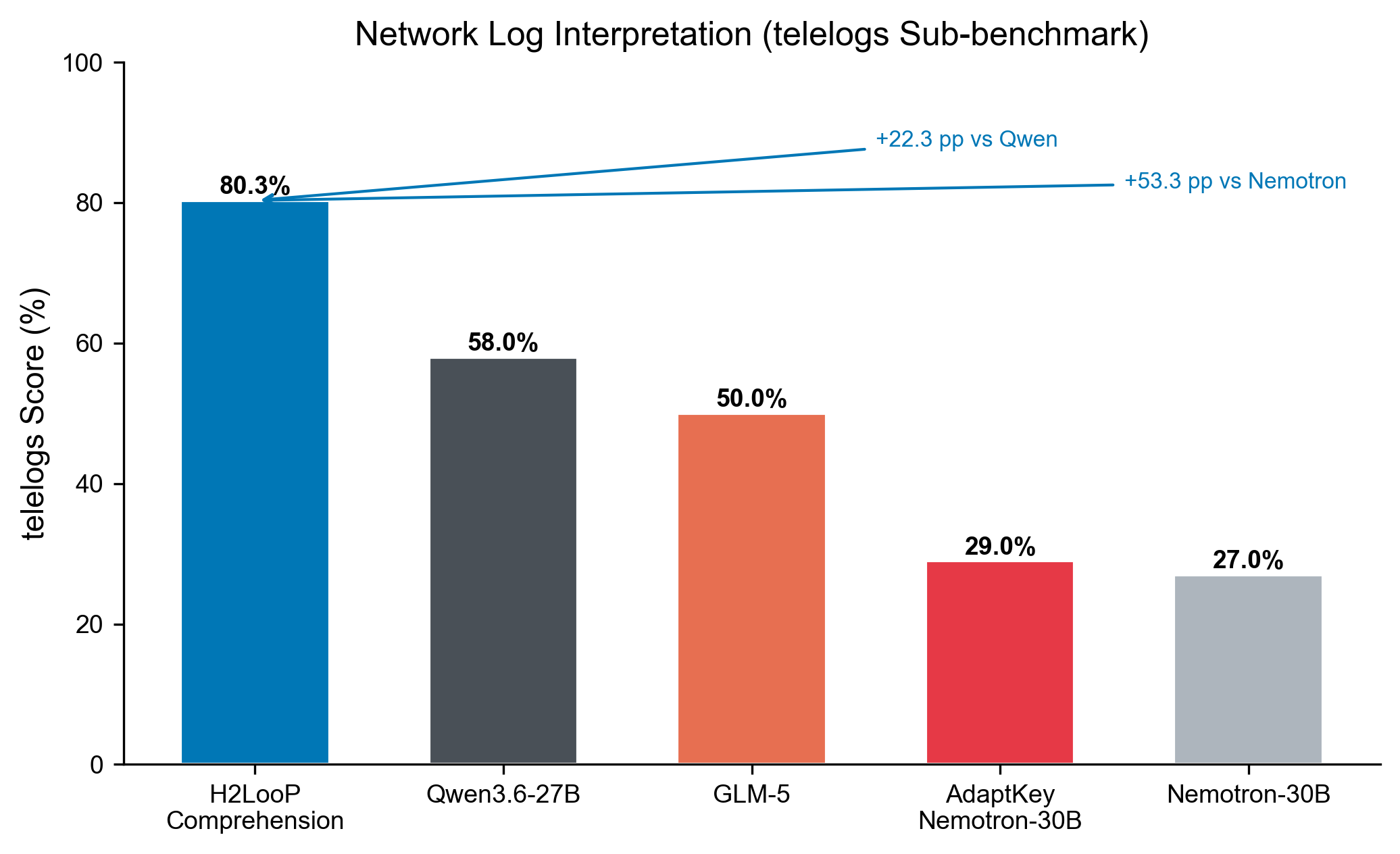}
    \caption{Network log interpretation (telelogs)-H2LooP dominance.}
    \label{fig:telelogs}
\end{figure}

\paragraph{Sub-benchmark Analysis.} Several sub-benchmarks merit detailed discussion:

\textbf{telelogs.} H2LooP Telecom Comprehension demonstrates particularly strong performance on network log interpretation (80.3\%), substantially outperforming Qwen3.6-27B (58.0\%, delta: +22.3~pp) and Nemotron-30B (27.0\%, delta: +53.3~pp). This result reflects the inclusion of network telemetry and log analysis data in our training corpus and validates the practical utility of the model for production debugging workflows.

\textbf{teleqna and teleqna-v2.} Both H2LooP variants achieve strong performance on the TeleQnA benchmarks (Comprehension: 86.6\%/86.2\%), outperforming Qwen3.6-27B, GLM-5, Nemotron-30B, and AdaptKey Nemotron-30B and approaching Gemini-class performance on standardized telecom Q\&A. On teleqna-v2 specifically, the Comprehension variant leads this comparison set at 86.2\%, demonstrating that the model has internalized precise 3GPP Release 17/18 specification details-exact parameter definitions, service operations, and message-level behavior-rather than relying on surface-level domain familiarity.

\textbf{telereason.} As the only open-ended benchmark in the OT-Lite suite, telereason is uniquely discriminative. It requires models to produce free-form operational answers spanning 5GC interface internals, fault diagnosis procedures, RAN parameter configurations, and multi-step protocol flows. The wide score distribution across models (6.0\% to 58.0\%) reflects the difficulty of generating technically precise responses without the scaffolding of multiple-choice options. H2LooP Telecom Comprehension achieves 51.3\%, with the benchmark specifically testing production-relevant capabilities: diagnosing faults, resolving handover issues, and configuring measurement setups with correct protocol messages and interface references.

\textbf{srsranbench and oranbench.} The Comprehension variant achieves 90.8\% on srsranbench and 88.9\% on oranbench, demonstrating deep understanding of both traditional and disaggregated RAN architectures.

Evaluation variance is characterized as within $\pm$5 percentage points across independent runs, verified over 20 repeated evaluations of the Comprehension variant.

\subsection{Independent Verification: Open Telco AI Leaderboard\textsuperscript{*}}
\label{sec:leaderboard}

Beyond our own OT-Lite Pass@3 evaluation harness (Table~\ref{tab:otlite_ranking}), we report H2LooP's standing on the community-run Open Telco AI Leaderboard, a third-party, blind evaluation of submitted models under a standardized harness independent of vendor self-reporting. H2LooP Telecom Comprehension ranks \textbf{5th overall} (AVG 74.8) among all evaluated models, open- and closed-source alike, at only 31B parameters-an order of magnitude smaller than several of the frontier proprietary systems it outperforms.

\begin{table}[H]
\centering
\caption{Open Telco AI Leaderboard-Overall Ranking Snapshot\textsuperscript{*\S\P}.}
\label{tab:leaderboard}
\resizebox{\textwidth}{!}{%
\begin{tabular}{clllc}
\toprule
\textbf{Rank} & \textbf{Model} & \textbf{Organization} & \textbf{Parameters} & \textbf{AVG} \\
\midrule
1 & OTel-2.0-LLM-31B-IT~\cite{otelatt}\textsuperscript{\P} & AT\&T & 31B & 90.3 \\
2 & OTel-LLM-8.3B-QnA~\cite{otelatt}\textsuperscript{\P} & AT\&T & 8.3B & 86.0 \\
3 & TeleLLM~\cite{telellm}\textsuperscript{\P} & China Telecom & -- & 75.8 \\
4 & gemini-3.1-pro-preview~\cite{gemini} & Google & Undisclosed & 75.6 \\
\textbf{5} & \textbf{H2LooP Telecom Comprehension} & \textbf{H2LooP} & \textbf{31B} & \textbf{74.8} \\
6 & gemini-3-pro-preview~\cite{gemini} & Google & Undisclosed & 74.7 \\
7 & LTM~\cite{ltm} & SoftBank & -- & 73.6 \\
8 & claude-opus-4.6~\cite{claude} & Anthropic & Undisclosed & 73.3 \\
9 & gpt-5~\cite{gpt5} & OpenAI & Undisclosed & 71.9 \\
10 & TSLAM-150B~\cite{tslam} & NetoAI & 150B & 71.6 \\
11 & gemini-3-flash-preview~\cite{gemini} & Google & Undisclosed & 70.4 \\
12 & claude-opus-4.5~\cite{claude} & Anthropic & Undisclosed & 69.6 \\
13 & kimi-k2.5~\cite{kimi} & Moonshot AI & Undisclosed & 69.4 \\
14 & o3~\cite{o3} & OpenAI & Undisclosed & 69.4 \\
15 & grok-4-fast~\cite{grok4} & xAI & Undisclosed & 68.4 \\
16 & claude-opus-4~\cite{claude} & Anthropic & Undisclosed & 68.2 \\
17 & o1~\cite{o1} & OpenAI & Undisclosed & 68.1 \\
18 & claude-opus-4.1~\cite{claude} & Anthropic & Undisclosed & 68.0 \\
19 & grok-4.1-fast~\cite{grok4} & xAI & Undisclosed & 67.1 \\
20 & claude-sonnet-4.5~\cite{claude} & Anthropic & Undisclosed & 66.0 \\
\bottomrule
\end{tabular}%
}
\vspace{0.3em}
\begin{minipage}{0.95\linewidth}
\small\textit{\P~We were unable to validate the reported numbers for OTel-2.0-LLM-31B-IT and OTel-LLM-8.3B-QnA (AT\&T); weights were not publicly available for TeleLLM (China Telecom) to independently verify its scores.}
\end{minipage}
\end{table}

At 31B parameters, H2LooP Telecom Comprehension outperforms, on this independently administered benchmark, thirteen frontier and near-frontier models from every major closed-source provider: \textbf{claude-opus-4.6}, \textbf{claude-opus-4.5}, \textbf{claude-opus-4.1}, \textbf{claude-opus-4}, and \textbf{claude-sonnet-4.5} (Anthropic); \textbf{gpt-5}, \textbf{o3}, and \textbf{o1} (OpenAI); \textbf{gemini-3-flash-preview} (Google); \textbf{grok-4-fast} and \textbf{grok-4.1-fast} (xAI); \textbf{kimi-k2.5} (Moonshot AI); and TSLAM-150B (NetoAI). This result is consistent with the parameter-efficiency argument developed in Section~\ref{sec:parameter_efficiency}: targeted domain specialization on curated telecom corpora closes-and on this leaderboard, surpasses-the gap to substantially larger, general-purpose frontier models on telecom-specific evaluation, under an evaluation protocol we do not control.

The absolute sub-benchmark scores on the Open Telco AI Leaderboard differ from our self-reported OT-Lite Pass@3 numbers (Table~\ref{tab:otlite_subbench}) because the leaderboard applies its own unweighted, single-attempt scoring harness rather than our internal weighted Pass@3 protocol; the two are not directly comparable in absolute terms, but the leaderboard's independent, blind evaluation corroborates H2LooP's competitive standing among frontier models.

\subsection{Telecom Coding: Autonomous Code Generation \& PR Resolution}
\label{sec:autonomous_pr}

H2LooP Telecom Agent is designed for end-to-end autonomous pull request generation on telecom codebases. Given a natural language description of a required change (bug fix, feature addition, or standards-compliance update), the model executes a three-stage pipeline:

\begin{enumerate}
    \item \textbf{File Localization:} Identifies relevant source files and code regions within the repository using learned structural priors about telecom project organization.
    \item \textbf{Edit Generation:} Produces syntactically correct, contextually appropriate code modifications conforming to project conventions and standards requirements.
    \item \textbf{Commit Structuring:} Organizes changes into coherent atomic commits with descriptive messages appropriate for code review.
\end{enumerate}

\subsubsection{Target Repositories}

We validate autonomous PR resolution capability across 11 production telecom repositories spanning the full software stack:

\begin{table}[H]
\centering
\caption{Target Repository Coverage.}
\label{tab:repos}
\resizebox{\textwidth}{!}{%
\begin{tabular}{llll}
\toprule
\textbf{Repository} & \textbf{Domain} & \textbf{Language} & \textbf{Description} \\
\midrule
srsRAN 4G~\cite{srsran4g} & RAN & C/C++ & 4G/LTE software radio suite \\
srsRAN Project~\cite{srsranproject} & RAN & C++ & 5G NR RAN implementation \\
Open5GS~\cite{open5gs} & Core & C & 5G core network (AMF, SMF, UPF, etc.) \\
free5GC~\cite{free5gc} & Core & Go & 5G core network implementation \\
OpenAirInterface5G~\cite{oai} & RAN/Core & C & 5G reference implementation \\
UERANSIM~\cite{ueransim} & Simulation & C++ & 5G UE and gNB simulator \\
NVIDIA Aerial~\cite{aerial} & RAN & C++/CUDA & CUDA-accelerated L1/L2 \\
gr-gsm~\cite{grgsm} & Legacy & C++/Python & GNU Radio GSM receiver \\
LTE-Cell-Scanner~\cite{ltescanner,ltescanner2} & Tools & C++ & LTE cell search and synchronization \\
RIS-Codes-Collection~\cite{riscodes} & Research & MATLAB/Python & RIS algorithm implementations \\
\bottomrule
\end{tabular}%
}
\end{table}

The repository set covers five distinct programming languages (C, C++, Go, Python, MATLAB), four telecom stack layers (physical, RAN, core, simulation), and project sizes ranging from 50K to 2M+ lines of code.

\subsubsection{Pipeline \& Metrics}

\begin{figure}[H]
    \centering
    \includegraphics[width=\linewidth]{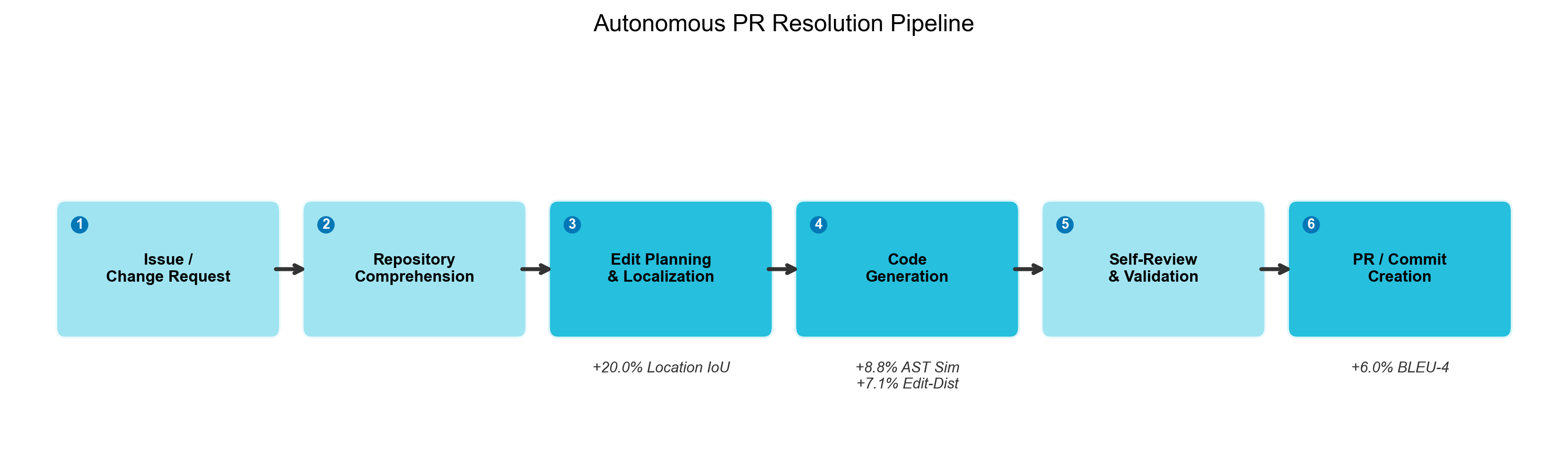}
    \caption{Autonomous PR resolution pipeline. Percentages indicate relative improvement of H2LooP Telecom Agent over the base model at each stage.}
    \label{fig:pipeline}
\end{figure}

Each pipeline stage corresponds to a measurable capability. We evaluate code generation quality on a held-out test split of telecom repository commits not observed during training, measuring four complementary dimensions:

\begin{itemize}
    \item \textbf{BLEU-4}~\cite{bleu}: N-gram precision measuring surface-level token overlap with reference edits.
    \item \textbf{AST Similarity}: Structural correctness via normalized tree edit distance between generated and reference ASTs.
    \item \textbf{Edit-Distance Score}: Normalized Levenshtein distance measuring character-level patch accuracy.
    \item \textbf{Location IoU}: Intersection over Union of predicted and ground-truth edit regions, measuring localization accuracy.
\end{itemize}

\begin{table}[H]
\centering
\caption{Pipeline Stage to Metric Mapping.}
\label{tab:pipeline_metrics}
\begin{tabular}{llc}
\toprule
\textbf{Pipeline Stage} & \textbf{Corresponding Metric} & \textbf{Score} \\
\midrule
Edit Planning (Localization) & Location IoU & 0.446 \\
Code Generation (Structural) & AST Similarity & 0.570 \\
Code Generation (Surface) & Edit-Distance Score & 0.519 \\
Overall Patch Quality & BLEU-4 & 0.338 \\
\bottomrule
\end{tabular}
\end{table}

\subsubsection{Quantitative Results}

\begin{table}[H]
\centering
\caption{Telecom Coding Benchmark Results.}
\label{tab:coding_benchmark}
\begin{tabular}{lcccc}
\toprule
\textbf{Metric} & \textbf{Base Model} & \textbf{H2LooP Agent} & \textbf{Rel. Impr.} & \textbf{AdaptKey~\cite{nemotron}} \\
\midrule
BLEU-4 & 0.3188 & \textbf{0.3380} & +6.0\% & 0.1360 \\
AST Similarity & 0.5240 & \textbf{0.5700} & +8.8\% & 0.2990 \\
Edit-Distance Score & 0.4848 & \textbf{0.5190} & +7.1\% & 0.3890 \\
Location IoU & 0.3717 & \textbf{0.4460} & +20.0\% & 0.2140 \\
\midrule
\textbf{Aggregate (mean)} & 0.4248 & \textbf{0.4683} & +10.2\% & 0.2595 \\
\bottomrule
\end{tabular}
\end{table}

\begin{figure}[H]
    \centering
    \includegraphics[width=0.85\linewidth]{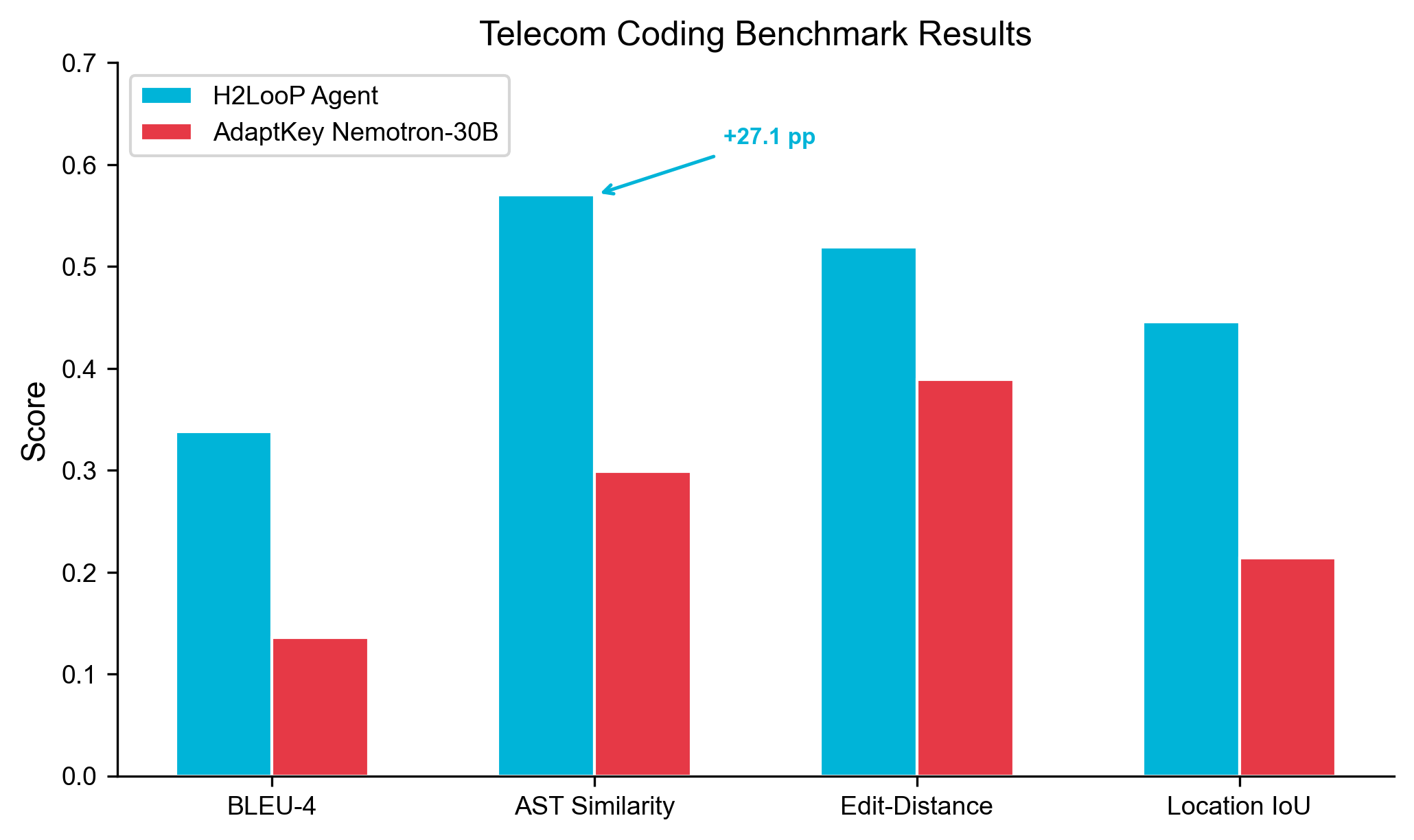}
    \caption{Telecom Coding Benchmark-H2LooP Agent vs.\ AdaptKey Nemotron-30B.}
    \label{fig:coding_benchmark}
\end{figure}

\begin{figure}[H]
    \centering
    \includegraphics[width=0.75\linewidth]{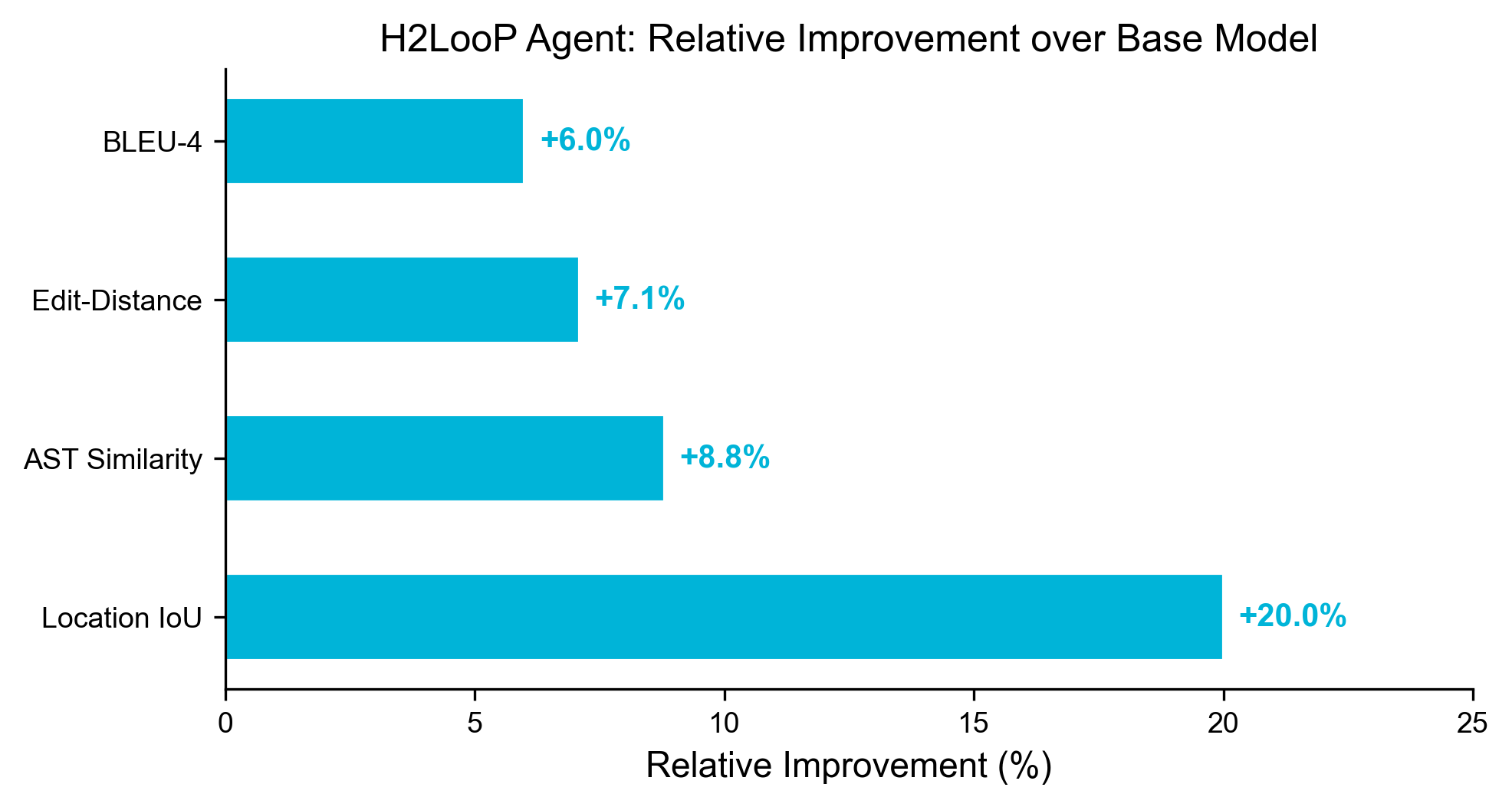}
    \caption{Relative improvement of H2LooP Telecom Agent over Base Model.}
    \label{fig:relative_improvement}
\end{figure}

H2LooP Telecom Agent improves over the base model across all four metrics, with the largest relative gain observed in Location IoU (+20.0\%). This metric captures the model's capacity to correctly identify the spatial location within a file where edits should be applied, a capability that requires deep structural understanding of telecom software architecture including module boundaries, function responsibilities, and protocol layer organization.

The absolute performance gap between H2LooP Telecom Agent and AdaptKey Nemotron-30B~\cite{nemotron} ranges from 13.0 to 27.1 percentage points across metrics (mean absolute delta: +20.9~pp), demonstrating that our fine-tuning approach substantially outperforms alternative domain adaptation strategies.

\begin{table}[H]
\centering
\caption{Absolute Performance Delta (H2LooP Agent vs.\ AdaptKey Nemotron-30B).}
\label{tab:absolute_delta}
\begin{tabular}{lccc}
\toprule
\textbf{Metric} & \textbf{H2LooP Agent} & \textbf{AdaptKey~\cite{nemotron}} & \textbf{Abs.\ Delta} \\
\midrule
BLEU-4 & 0.3380 & 0.1360 & +0.2020 \\
AST Similarity & 0.5700 & 0.2990 & +0.2710 \\
Edit-Distance Score & 0.5190 & 0.3890 & +0.1300 \\
Location IoU & 0.4460 & 0.2140 & +0.2320 \\
\bottomrule
\end{tabular}
\end{table}

\subsubsection{Qualitative Analysis}

We observe the following qualitative properties of generated PRs across target repositories:

\begin{itemize}
    \item \textbf{Standards compliance:} Generated code correctly references 3GPP procedure definitions and implements protocol state transitions consistent with specification requirements.
    \item \textbf{Project convention adherence:} The model adapts to per-repository coding style including naming conventions, error handling patterns, and logging frameworks.
    \item \textbf{Multi-file coherence:} Changes spanning multiple files maintain consistency across header declarations, implementation bodies, and test modifications.
    \item \textbf{Build system awareness:} Generated edits account for build dependencies, CMake configurations, and conditional compilation directives.
\end{itemize}

\subsection{General Capability Preservation}

A critical requirement for domain-specialized models is retention of general capabilities. We evaluate on two axes: general knowledge (MMLU~\cite{mmlu}) and multi-turn function calling (BFCL v3~\cite{bfcl}).

\subsubsection{MMLU}

\begin{table}[H]
\centering
\caption{MMLU Performance (Anti-Forgetting Evaluation).}
\label{tab:mmlu}
\begin{tabular}{lcc}
\toprule
\textbf{Model} & \textbf{MMLU} & \textbf{$\Delta$ vs.\ Base} \\
\midrule
Base Model & 74.0\% & -- \\
\textbf{H2LooP Telecom Agent} & \textbf{74.0\%} & \textbf{0.0~pp} \\
Qwen3.6-27B~\cite{qwen3} & 54.0\% & N/A \\
\bottomrule
\end{tabular}
\end{table}

H2LooP Telecom Agent achieves identical MMLU performance to the base model (74.0\%), confirming zero catastrophic forgetting of general knowledge. Our result demonstrates that LoRA-based domain fine-tuning can achieve strong specialization without compromising the parameter space allocated to general knowledge.

\subsubsection{Multi-Turn Function Calling: BFCL v3}

Agentic deployment requires robust multi-turn function calling capability. We evaluate on the Berkeley Function Calling Leaderboard v4 (BFCL v3)~\cite{bfcl} using the OpenAI-compatible tool calling API across four multi-turn scenario categories.

\begin{table}[H]
\centering
\caption{BFCL v3 Multi-Turn Function Calling Results.}
\label{tab:bfcl}
\begin{tabular}{lccccc}
\toprule
\textbf{Model} & \textbf{base} & \textbf{long\_ctx} & \textbf{miss\_func} & \textbf{miss\_param} & \textbf{Mean} \\
\midrule
Base Model & 79.0\% & 71.5\% & 44.5\% & 50.5\% & 61.4\% \\
H2LooP Telecom Agent & 79.0\% & 71.5\% & 42.0\% & 50.0\% & 60.6\% \\
AdaptKey Nemotron-30B~\cite{nemotron} & 24.5\% & 14.0\% & 8.0\% & 12.0\% & 14.6\% \\
\bottomrule
\end{tabular}
\end{table}

\begin{table}[H]
\centering
\caption{BFCL v3 Capability Retention Rate (H2LooP Agent / Base Model).}
\label{tab:bfcl_retention}
\begin{tabular}{lc}
\toprule
\textbf{Category} & \textbf{Retention Rate} \\
\midrule
multi\_turn\_base & 100.0\% \\
multi\_turn\_long\_context & 100.0\% \\
multi\_turn\_miss\_func & 94.4\% \\
multi\_turn\_miss\_param & 99.0\% \\
\midrule
\textbf{Mean Retention} & \textbf{98.4\%} \\
\bottomrule
\end{tabular}
\end{table}

\begin{figure}[H]
    \centering
    \includegraphics[width=0.85\linewidth]{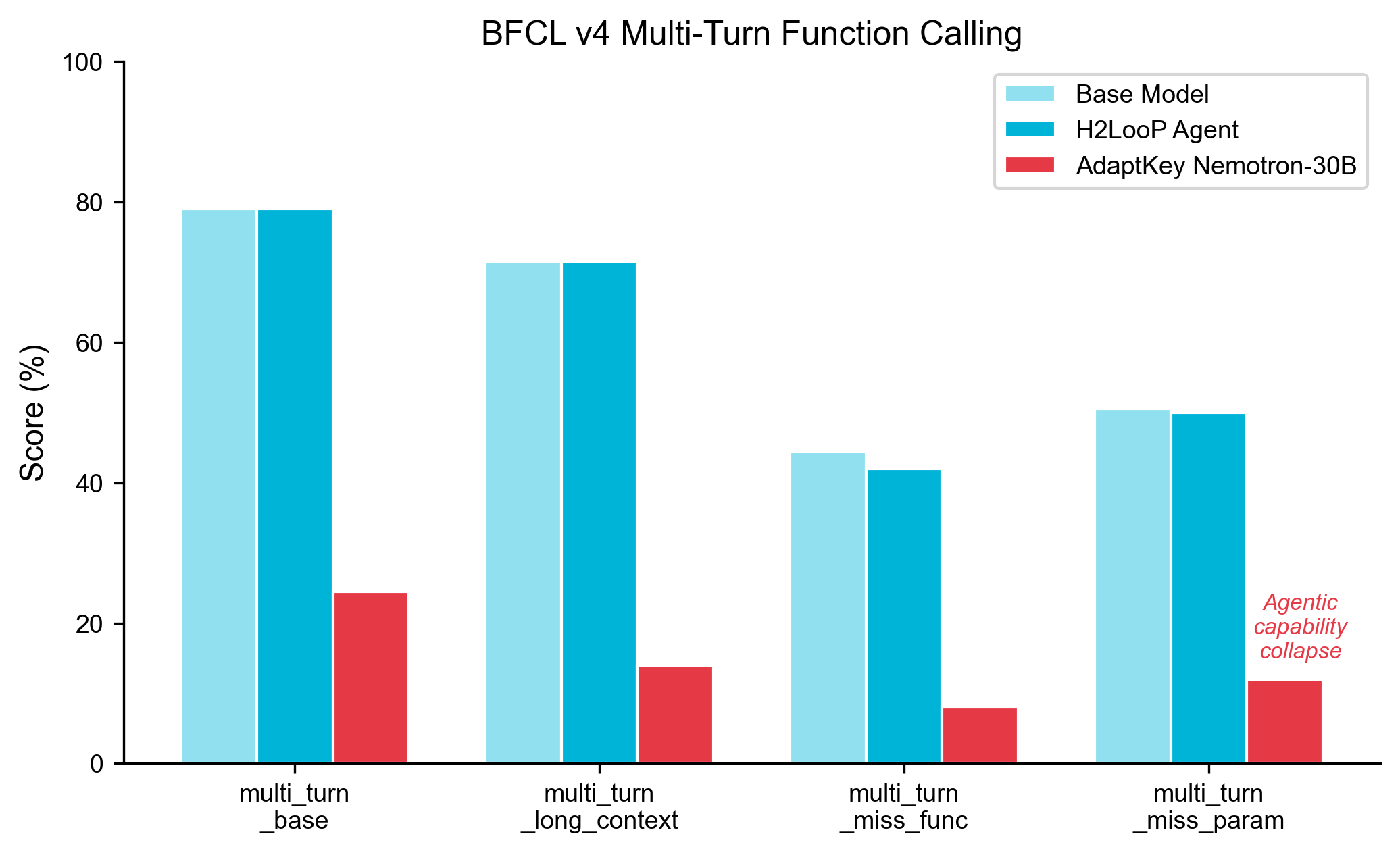}
    \caption{BFCL v3 multi-turn function calling-capability preservation vs.\ collapse.}
    \label{fig:bfcl}
\end{figure}

\begin{figure}[H]
    \centering
    \includegraphics[width=0.75\linewidth]{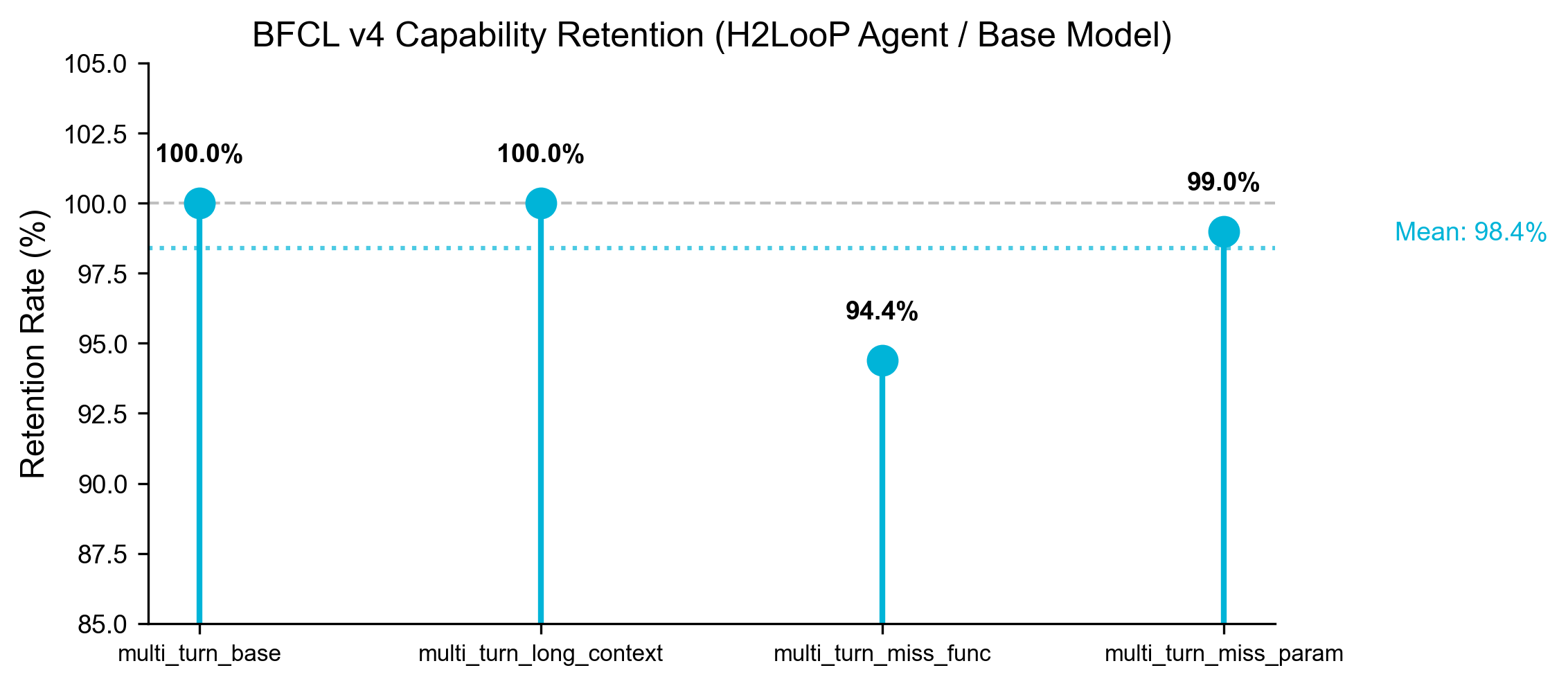}
    \caption{BFCL v3 capability retention rate (H2LooP Agent / Base Model).}
    \label{fig:retention}
\end{figure}

H2LooP Telecom Agent achieves a mean capability retention rate of 98.4\% relative to the base model across all BFCL v3 categories, with identical performance on the two primary categories (multi\_turn\_base: 79.0\%, multi\_turn\_long\_context: 71.5\%). This stands in contrast to AdaptKey Nemotron-30B~\cite{nemotron}, which retains only 23.8\% of base-class function calling performance (mean: 14.6\% vs.\ reported base capability), rendering it unsuitable for agentic deployment.

\subsection{Composite Performance Analysis}

To provide a holistic view of model capability across all evaluation axes, we compute a normalized composite score aggregating telecom coding, comprehension, function calling, and general knowledge metrics.

\begin{table}[H]
\centering
\caption{Normalized Composite Score (0--100 scale).}
\label{tab:composite}
\begin{tabular}{lccccc}
\toprule
\textbf{Model} & \textbf{Telecom Coding} & \textbf{OT-Lite} & \textbf{BFCL v3} & \textbf{MMLU} & \textbf{Composite} \\
\midrule
H2LooP Telecom Agent & 81.3 & 78.8 & 76.6 & 74.0 & \textbf{77.7} \\
Base Model & 73.7 & 79.3 & 77.5 & 74.0 & 76.1 \\
Qwen3.6-27B~\cite{qwen3} & -- & 80.0 & -- & 54.0 & -- \\
AdaptKey Nemotron-30B~\cite{nemotron} & 45.0 & 68.9 & 18.5 & 84.0 & 54.1 \\
Nemotron-30B~\cite{nemotron} & -- & 69.1 & -- & -- & -- \\
\bottomrule
\end{tabular}
\end{table}

Normalization: Telecom Coding is the mean of four metrics scaled to 0--100; BFCL v3 is the mean across four categories; OT-Lite and MMLU are reported directly. The composite is the unweighted arithmetic mean of available metrics.

\begin{figure}[H]
    \centering
    \includegraphics[width=0.7\linewidth]{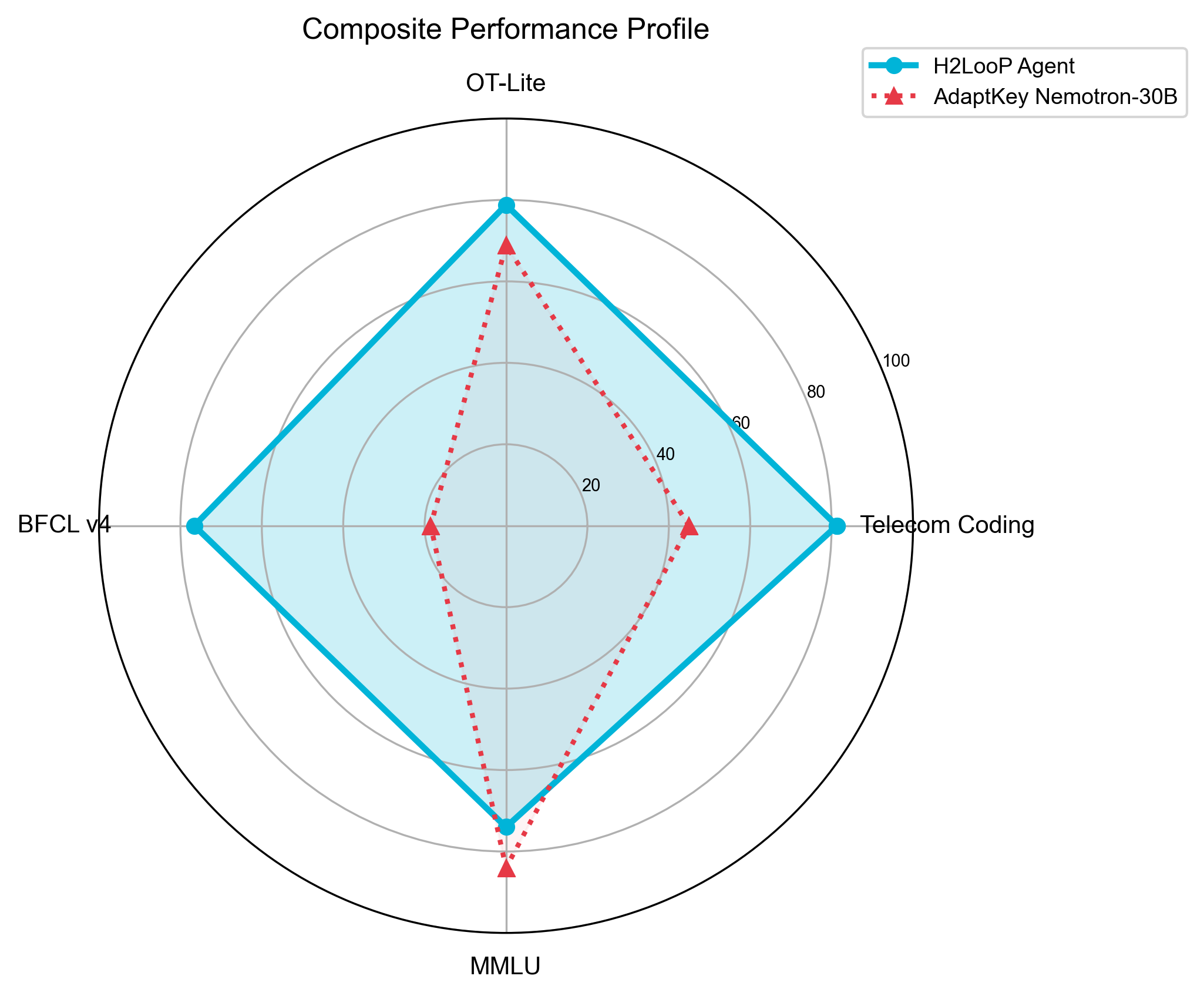}
    \caption{Composite performance profile-balanced multi-capability.}
    \label{fig:composite_radar}
\end{figure}

\section{Discussion}

\subsection{Specialization Without Catastrophic Forgetting}

A central finding of this work is that telecommunications domain specialization can be achieved without measurable degradation of general capabilities. H2LooP Telecom Agent maintains identical MMLU performance (74.0\%) and near-identical BFCL v3 function calling performance (98.4\% retention) relative to the base model, while gaining substantial telecom-specific improvements across coding and comprehension evaluations. This result validates the hypothesis that LoRA-based fine-tuning can inject domain knowledge into underutilized capacity within the model's parameter space without overwriting general-purpose representations.

This finding contrasts with AdaptKey Nemotron-30B~\cite{nemotron}, where architectural modifications for domain adaptation induce catastrophic degradation of function calling capability (14.6\% mean BFCL v3 vs.\ 60.6\% for H2LooP Telecom Agent), demonstrating that not all adaptation strategies preserve the multi-capability balance required for agentic deployment.

\subsection{Domain Comprehension as a Foundation for Code Quality}

The synergy between comprehension and coding training is evidenced by the Agent variant's performance profile. Despite being optimized for code generation, it maintains 78.8\% on OT-Lite (only 3.0~pp below the pure Comprehension variant). We hypothesize that deep telecom domain understanding directly improves code generation quality by enabling the model to reason about the semantic intent behind code changes, not merely their syntactic form. The telelogs sub-benchmark performance (76.0\% for Agent vs.\ 58.0\% for Qwen3.6-27B) exemplifies this: understanding network log semantics is prerequisite to generating correct log parsing and analysis code.

\subsection{Parameter Efficiency of Domain Specialization}
\label{sec:parameter_efficiency}

The observation that our model surpasses Qwen3-480B~\cite{qwen3} (72.9\% OT-Lite) by 8.9~pp despite having substantially fewer parameters carries significant implications for telecom deployment. In network operations environments where inference latency and compute cost are constrained, a specialized model achieving 81.8\% accuracy is substantially more deployable than a 480B general-purpose model achieving 72.9\%. The parameter efficiency ratio (performance per billion parameters) of our approach exceeds general scaling by an order of magnitude on domain-specific evaluation.

\subsection{The Agentic Capability Prerequisite}

The BFCL v3 evaluation reveals a critical requirement for domain-specialized models intended for autonomous operation: function calling capability must be explicitly preserved through the fine-tuning process. AdaptKey Nemotron-30B~\cite{nemotron} achieves acceptable MMLU (84.0\%) and moderate OT-Lite (68.9\%) scores but is rendered entirely non-functional as an agent due to 14.6\% mean multi-turn function calling performance. Our results demonstrate that careful training data composition and fine-tuning methodology can maintain agentic capabilities at 98.4\% of base model performance while simultaneously achieving domain specialization.

\begin{figure}[H]
    \centering
    \includegraphics[width=0.75\linewidth]{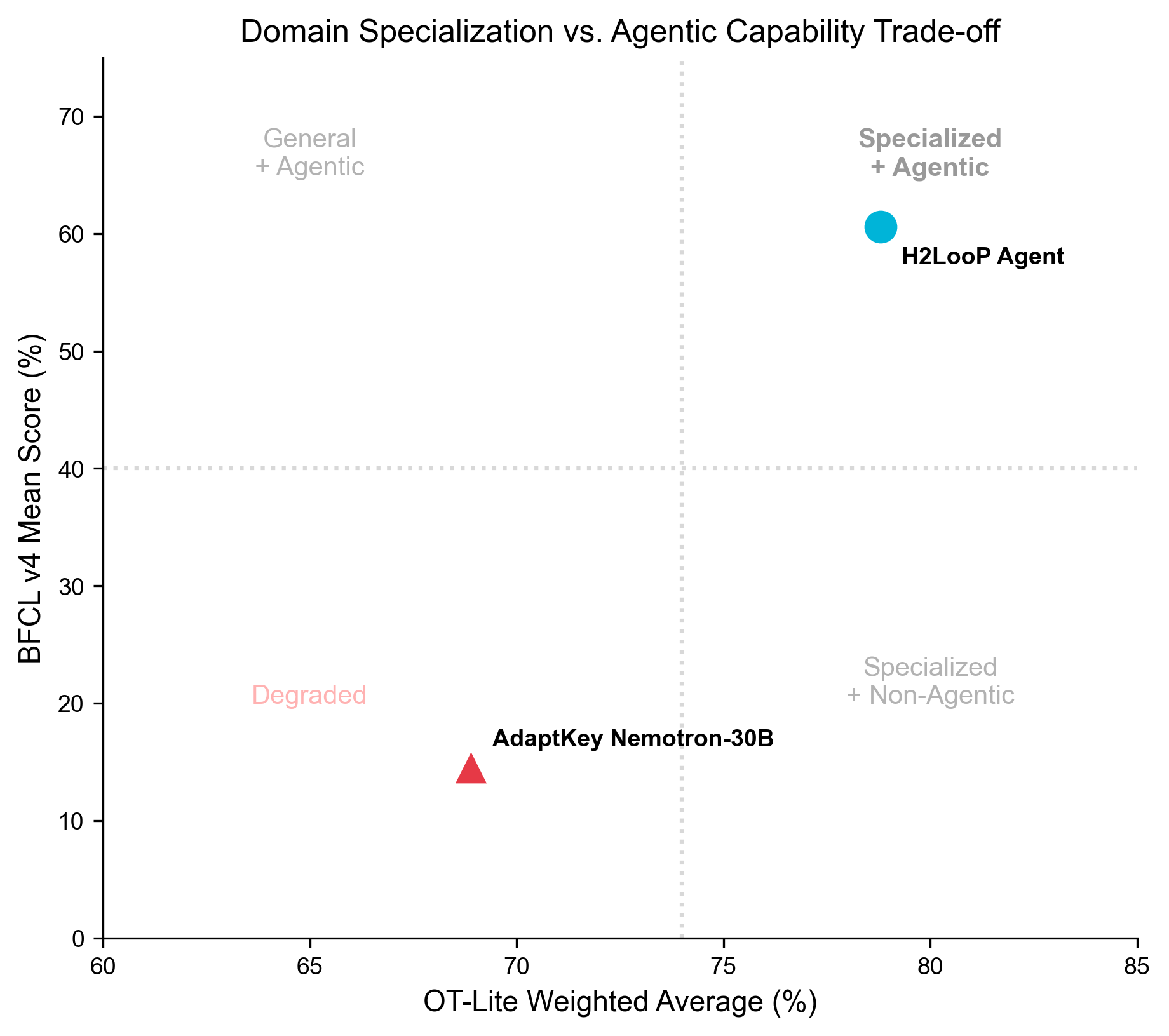}
    \caption{Domain specialization vs.\ agentic capability trade-off.}
    \label{fig:tradeoff}
\end{figure}

\section{Limitations and Future Work}

\textbf{Benchmark scale.} Our telecom coding benchmark, while derived from authentic repository commits, is limited in total sample count. Future work will expand evaluation to end-to-end PR acceptance rates on live repositories with automated CI/CD validation.

\textbf{Temporal currency.} Telecom standards evolve continuously with new 3GPP releases. Maintaining model currency requires periodic re-training on updated specification corpora, and we plan to establish a continuous fine-tuning pipeline.

\textbf{End-to-end deployment metrics.} While component metrics (Location IoU, AST Similarity) are reported, integration-level metrics such as PR merge rate, first-pass review approval rate, and time-to-resolution in production CI/CD environments represent important future evaluation targets.

\textbf{Model scale.} The current release is parameter-constrained. Investigation of scaling laws for domain-specialized telecom models, including mixture-of-experts architectures, constitutes a natural extension.

\section{Conclusion}

We have presented H2LooP Telecom Model v1, a domain-specialized large language model achieving strong performance on telecommunications comprehension and coding benchmarks, outperforming frontier closed-source models such as GPT-5 and Claude Opus on independent leaderboard evaluation. The model demonstrates three key properties: (1)~superior domain performance relative to general-purpose models of equivalent and substantially larger scale, (2)~zero catastrophic forgetting of general knowledge and agentic capabilities, and (3)~practical autonomous PR resolution capability across diverse production telecom repositories. Our results establish that targeted fine-tuning on curated domain corpora represents a highly parameter-efficient approach to telecom AI, outperforming naive scaling by an order of magnitude on domain-specific evaluation while preserving the multi-capability balance required for agentic deployment in production software engineering workflows.

\section*{Notes}

\begin{enumerate}
\item[\textsuperscript{*}] Independent leaderboard results reported in Section~\ref{sec:leaderboard} reflect the official Open Telco AI Leaderboard standing as of September 1, 2026; rankings are subject to change as new model versions are submitted.

\item[\textsuperscript{\dag}] TSLAM-150B (NetoAI), reported in Table~\ref{tab:otlite_subbench} and Figure~\ref{fig:radar_otlite}, is not part of our self-reported baseline set; scores are sourced directly from the official Open Telco AI Leaderboard, which evaluates only 7 of the 10 OT-Lite sub-benchmarks (sixg-bench, teleqna-v2, and telereason are marked ``--''). As this uses a different scoring harness than our internal weighted Pass@3 protocol, comparisons should be read as directional rather than exact.

\item[\textsuperscript{\ddag}] Reported scores in Table~\ref{tab:otlite_ranking} reflect the model versions publicly available at the time of measurement; model availability and reported capabilities are subject to change as vendors release updates.

\item[\textsuperscript{\S}] Reported ranks and scores in Table~\ref{tab:leaderboard} reflect a snapshot of the leaderboard at the time of measurement; standings are subject to change as new model versions are submitted.
\end{enumerate}

\end{document}